\documentclass[10pt,twocolumn,letterpaper]{article}

\usepackage{cvpr}
\usepackage{times}
\usepackage{epsfig}
\usepackage{graphicx}
\usepackage{amsmath}
\usepackage{amssymb}

\usepackage{subfigure}
\usepackage{algorithm}
\usepackage{algorithmic}
\usepackage{placeins}

\usepackage[pagebackref=true,breaklinks=true,letterpaper=true,colorlinks,bookmarks=false]{hyperref}

\cvprfinalcopy 

\def\cvprPaperID{1693} 

\ifcvprfinal\fi
\begin{document}

\title{A Deep Learning Based Fast Image Saliency Detection Algorithm}

\author{Hengyue Pan\\
York University\\
4700 Keele Street, Toronto, Ontario, CA\\
{\tt\small panhy@cse.yorku.ca}\\
\and
Hui Jiang\\
York University\\
4700 Keele Street, Toronto, Ontario, CA\\
{\tt\small hj@cse.yorku.ca}
}

\maketitle

\begin{abstract}
In this paper, we propose a fast deep learning method for object saliency detection using convolutional neural networks. In our approach, we use a gradient descent method to iteratively modify the input images based on the pixel-wise gradients to reduce a pre-defined cost function, which is defined to measure the class-specific objectness and clamp the class-irrelevant outputs to maintain image background.  The pixel-wise gradients can be efficiently computed using the back-propagation algorithm. We further apply SLIC superpixels and LAB color based low level saliency features to smooth and refine the gradients. Our methods are quite computationally efficient, much faster than other deep learning based saliency methods. Experimental results on two benchmark tasks, namely Pascal VOC 2012 and MSRA10k, have shown that our proposed methods can generate high-quality salience maps, at least comparable with many slow and complicated deep learning methods. Comparing with the pure low-level methods, our approach excels in handling many difficult images, which contain complex background, highly-variable salient objects, multiple objects, and/or very small salient objects.
\end{abstract}

\section{Introduction}
\label{sec_introduction}

In the past few years, deep convolutional neural networks (DCNNs) \cite{lecun1995convolutional} have achieved the state-of-the-art performance in many computer vision tasks, starting from image recognition \cite{NIPS2012_AlexNet,GoogleLeNet_2014,VGG_2015} and object localization \cite{overfeat_2014} and more recently extending to object detection and semantic image segmentation \cite{r-cnn_2014,hariharan2014simultaneous}.
These successes are largely attributed to the capacity that large-scale DCNNs can effectively learn end-to-end from a large amount of labelled images in a supervised learning mode.

In this paper, we consider to apply the popular deep learning techniques to another computer vision problem, namely object saliency detection.
The saliency detection attempts to locate the objects that have the most interests in an image, where human may also pay more attention \cite{liu2011learning}.
The main goal of the saliency detection is to compute a saliency map that topographically represents the level of saliency for visual attention \cite{BooleanMap-2013}. For each pixel in an image, the saliency map can provide how likely this pixel belongs to the salient objects \cite{borji2012salient}. Computing such saliency maps has recently raised a great amount of research interest \cite{visual-model-2013}.
The computed saliency maps have been shown to be beneficial to various vision tasks, such as
image segmentation \cite{cheng2011global}, object recognition  and visual tracking.
The saliency detection has been extensively studied in computer vision, and a variety of methods have been proposed to generate the saliency maps for images. Under the assumption that the salient objects probably are the parts that significantly differ from their surroundings, most of the existing methods use low-level image features to detect saliency regions based on the criteria related to color contrast, rarity and symmetry of image patches \cite{cheng2011global,liu2011learning,riche2012rare,borji2012salient,fu2013superpixel}. In some cases, the global topological cues may be leveraged to refine the perceptual saliency maps \cite{harel2006graph,BooleanMap-2013,li2010probabilistic}. In these methods, the saliency is normally measured based on different mathematical models, including decision theoretic models, Bayesian models, information theoretic models, graphical models, and spectral analysis models \cite{visual-model-2013}.

Different from the previous low level methods, we propose a novel deep learning method for the object saliency detection based on the powerful DCNNs. As shown in \cite{NIPS2012_AlexNet,GoogleLeNet_2014,VGG_2015}, relying on a pre-trained classification DCNN, we can achieve a fairly high accuracy in object category recognition for many real-world images. Even though DCNNs can recognize what kind of objects are contained in an image, it is not straightforward for them to precisely locate the recognized objects in the image. In \cite{overfeat_2014,r-cnn_2014,hariharan2014simultaneous}, some
rather complicated and time-consuming post-processing stages are needed to detect and locate the objects for semantic image segmentation. In \cite{zhao2015saliency}, two DCNNs are applied to generate superpixel based global saliency features and local saliency features, which should be combined for the final saliency maps.

In this work, we propose a much simpler and more computationally efficient method to generate a class-specific object saliency map directly from the classification DCNN model. In our approach, we use a gradient descent (GD) method to iteratively modify each input image based on the refined pixel-wise gradients to reduce a pre-defined cost function, which is defined to measure the class-specific objectness and clamp the class-irrelevant outputs to maintain image background.  The gradients with respect to all image pixels can be efficiently computed using the back-propagation algorithm for DCNNs. After the back-propagation procedure, the discrepancy between the modified image and the original one is calculated as the raw saliency map for this image. The raw saliency maps are smoothed by using SLIC \cite{achanta2012slic} superpixel maps and refined by using low level saliency features. Since we only need to run a very small number of GD iterations in the saliency detection, our methods are extremely computationally efficient (average processing time for one image in one GPU is around 0.45 second).

Experimental results on two databases, namely Pascal VOC 2012 \cite{Everingham10} and MSRA10k \cite{SalObjSurvey}, have shown that our proposed methods can generate high-quality salience maps, at least comparable with many slow and complicated deep learning methods.   On the other hand, comparing with the traditional low-level methods, our  approach excels on many difficult images, containing complex background, highly-variable salient objects, multiple objects, and/or very small objects.

\section{Related Work}
\label{sec_related_work}

In the literature, the previous saliency detection methods mostly adopt the well-known bottom-up strategy \cite{cheng2011global,liu2011learning,riche2012rare,borji2012salient}. They relies on the local image features derived from patches to detect contrast, rarity and symmetry to identify the salient objects in an image. Meanwhile, some other methods have been proposed to take into account some global information or prior knowledge to screen the local features. For example,  in \cite{BooleanMap-2013}, a boolean map is created to represent global topological cues in an image, which in turn is used to guide the generation of saliency maps.
In \cite{li2010probabilistic}, the visual saliency algorithm considers the prior information and the local features simultaneously in a probabilistic model.The algorithm defines task-related components as the prior information to help the feature selection procedure. In \cite{cheng2011global}, a region contrast based image saliency method is proposed to generate the saliency maps, in which the global contrast differences are evaluated as the main saliency features. In \cite{fu2013superpixel}, the SLIC superpixels are used as the unit to generate the global contrast based saliency maps, and an average ground truth prior is introduced to eliminate some false positives. This research also takes color distribution information into account to further refine the saliency maps.
The traditional saliency detection methods normally work well for the images containing simple dominant foreground objects in homogenous backgrounds. However, they are usually not robust enough to handle images containing complex scenes \cite{li2014visual}, such as the relatively small objects in heterogenous backgrounds .


Recently, some deep learning techniques have been proposed for image saliency detection and semantic image segmentation \cite{overfeat_2014,r-cnn_2014,hariharan2014simultaneous,zhao2015saliency}. These methods typically use DCNNs to examine a large number of region proposals from other algorithms, and use the features generated by DCNNs along with other post-stage classifiers to localize the target objects. And currently more and more methods tend to directly generate pixel-wise saliency maps or segmentation \cite{hariharan2014simultaneous}. For example, in \cite{zhao2015saliency}, two DCNNs are applied to model the global context and local context for each superpixel in the input images, and the two levels of context are finally combined to generate the pixel-wise multi-context saliency maps.

In this paper, instead of directly generating the high-level semantic saliency maps from DCNNs, we propose to use DCNNs to generate middle-level saliency maps in a very efficient way, which may be fed to other traditional computer vision algorithms for various vision tasks, such as semantic segmentation, video tracking, etc.
The work in \cite{Oxford-cnn-2014} is the most relevant to the work in this paper. In \cite{Oxford-cnn-2014}, the authors have borrowed the idea of explanation vectors in \cite{Baehrens-2010} to generate a static pixel-wise gradient vector of the network learning objective function, and use it as a saliency map. In our work, an iterative gradient descent method is proposed to generate more reliable and robust saliency maps. More importantly, we introduce a new cost function for the back-propagation and apply SLIC superpixel maps and low level saliency features to refine the gradients for better saliency performance. 

The structure of the rest of this paper are listed below:  Section~\ref{sec_sal_seg} defines our proposed saliency and segmentation algorithm; Section~\ref{sec_experiments} shows experiment results of different databases and compare with the state-of-the-art method; finally Section~\ref{sec_conc} provides the conclusion.

\begin{figure*}[t]
\begin{center}
    \includegraphics[width=0.9\linewidth]{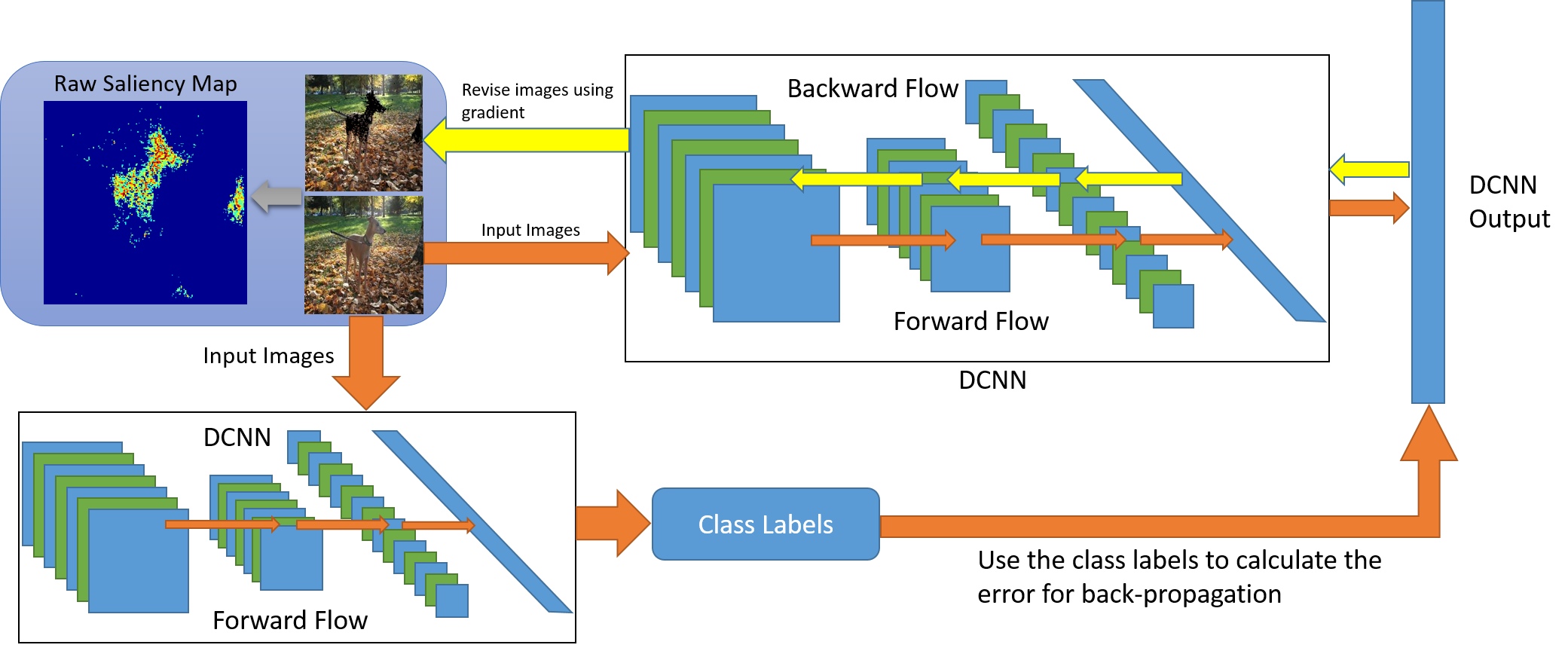}
\end{center}
\caption{The proposed method to generate the object-specific saliency maps directly from DCNNs.}
\label{Fig:Overview_saliency}
\end{figure*}

\section{Our Approach for Object Saliency Detection}
\label{sec_sal_seg}

In this section we will consider the main idea of our DCNN based saliency detection method, and also discuss how to smooth and refine the raw saliency map for better performance.

\subsection{Backpropagating and partially clamping DCNNs to generate raw saliency maps}
\label{subsec_DCNNsal}

As we have known, DCNNs can automatically learn all sorts of features from a large amount of labelled images, and a well-trained DCNN can achieve a very good classification accuracy in recognizing objects in images. In this work, based on the idea of explanation vectors in \cite{Baehrens-2010}, we argue that the classification DCNNs themselves may have learned enough features and information to generate good object saliency for the images. Extending a preliminary study in \cite{Oxford-cnn-2014}, we explore a novel method to generate the saliency maps directly from DCNNs. The key idea of our approaches is shown in Figure~\ref{Fig:Overview_saliency}. After an input image is recognized by a DCNN as containing one particular object, if we can modify the input image in such a way that the DCNN no longer recognizes the object from it and meanwhile attempts to maintain image background as much as possible, the discrepancy between the modified image and the original one may serve as a good saliency map for the recognized object. In this paper, we propose to use a gradient descent (GD) method to iteratively modify the input image based on the pixel-wise gradients to reduce a cost function formulated in the output layer of the DCNN. The proposed cost function is defined to measure the class-specific objectness. The cost function is reduced under the constraint that all class-irrelevant DCNN outputs are clamped to the original values. The image is modified by the gradients computed by applying the back-propagation procedure all the way to the input layer. In this way, the underlying object may be erased from the image while the irrelevant background may be largely retained.

First of all, we simply train a regular DCNN for the image classification. After the DCNN is learned, we may apply our saliency detection method to generate the class-specific object saliency map. For each input image $X$, we firstly use the pre-trained classification DCNN to generate its class label, denoted as $l$, as in a normal classification step. Meanwhile, we obtain the DCNN outputs prior to the final softmax layer, denoted as $\{ o_k \; | \; k=1,\cdots,N \}$.
Apparently, $o_l$ achieves the maximum value (due to the image is recognized as $l$). Here, we assume that the DCNN output $o_l$ is mainly relevant to the underlying object in the image while the remaining DCNN outputs $ \{o_k \; | \; k \neq l \}$ are more relevant to the image background excluding the underlying object.
Under this assumption, we propose a procedure to modify the image to reduce the $l$-th output of the DCNN as much as possible and meanwhile clamp the other outputs to their original values ${o_k}$. We further denote the output nodes (prior to softmax) of the DCNN in the saliency generation procedure as $\{ a_i \; | \; i=1,\cdots,N \}$. Therefore, for the image $X$,  we attempt modify $X$ to reduce the corresponding largest DCNN output, i.e. $a_l$,
subject to the constraint that all remaining DCNN outputs are clamped to their initial values:
$$
a_k = o_k   \;\;\; ( k=1,\cdots,N \; \mbox{and} \; k\neq l).
$$

Next, we propose to cast the above constraints as penalty terms to construct the following cost function:
\begin{equation}\label{eq-costfunction-CNN1}
{\cal F} (X | l ) =  a_l + \frac{\gamma}{2} \sum_{k^ \neq l} (a_{k} - o_{k})^2
\end{equation}
where $\gamma$ is a hyperparameter to balance the contribution from the constraints.
In this way, we have converted the original constrained optimization problem into an unconstrained problem, which can be easily minimized by gradient descent (GD) methods.



Obviously, this cost function is constructed based on the assumption that the recognized $l$-th output of the DCNN, i.e. $a_l$, corresponds to the foreground area in the input image while the remaining outputs of DCNN are more relevant to the image background. Therefore, if we modify the image $X$ to reduce the above cost function and hopefully the underlying object (belonging to class $l$) will be removed as the consequence due to that fact that $a_l$ is reduced significantly, but the background remains largely unchanged due to the rest DCNN outputs are clamped in this procedure. In this paper, we propose to use an iterative GD procedure to modify $X$ as follows:

\begin{equation}\label{eq-SGD-saliency}
X^{(t+1)} \gets X^{(t)} - \epsilon \cdot \max \left(\frac{\partial {\cal F}(X | l ) }{\partial X} \Big|_{X=X^{(t)}}, 0 \right)
\end{equation}
where $\epsilon$ is the learning rate, and we floor all negative gradients in the GD updates. We have observed in our experiments that the cost function ${\cal F} (X | l )$ can be significantly reduced by running only a small number of updates (typically 5-10 iterations) for each image, which guarantees the efficiency of the proposed method.

We can easily compute the above gradients using the standard back-propagation algorithm. Based on the cost function ${\cal F} (X | l )$ in Eq.(\ref{eq-costfunction-CNN1}), we can derive the error signals in the output layer,
$e_i = \frac{\partial {\cal F} (X | l ) }{ \partial a_i}$  ($i=1,\cdots, N$), as follows:
\begin{equation}
e_i =
    \begin{cases}
    \gamma (a_i - o_i)& \text{if $i \neq l$}, \\
    1 & \text{if $i = l$}.
    \end{cases}
\end{equation}


These error signals are back-propagated all the way to the input layer to derive the above gradient, $\frac{\partial {\cal F} (X | l ) }{\partial X}$, for saliency detection.

\begin{figure}[tb]
\begin{minipage}{0.23\linewidth}
    \centerline{\includegraphics[width=2.00cm]{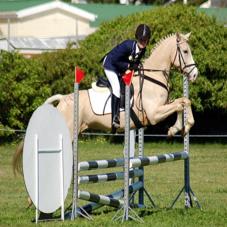}}
\end{minipage}
\hfill
\begin{minipage}{0.23\linewidth}
    \centerline{\includegraphics[width=2.00cm]{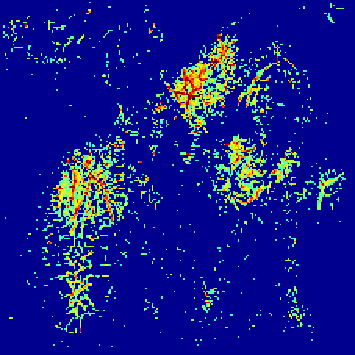}}
\end{minipage}
\hfill
\begin{minipage}{0.23\linewidth}
    \centerline{\includegraphics[width=2.00cm]{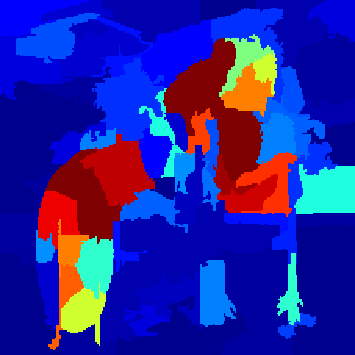}}
\end{minipage}
\hfill
\begin{minipage}{0.23\linewidth}
    \centerline{\includegraphics[width=2.00cm]{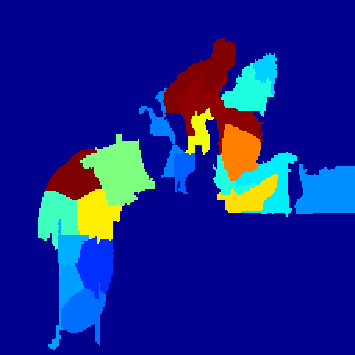}}
\end{minipage}
\vfill

\begin{minipage}{0.23\linewidth}
    \centerline{\includegraphics[width=2.00cm]{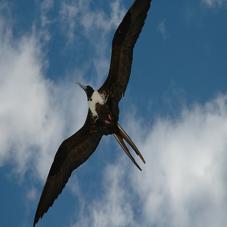}}
\end{minipage}
\hfill
\begin{minipage}{0.23\linewidth}
    \centerline{\includegraphics[width=2.00cm]{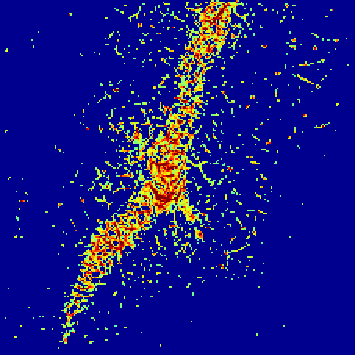}}
\end{minipage}
\hfill
\begin{minipage}{0.23\linewidth}
    \centerline{\includegraphics[width=2.00cm]{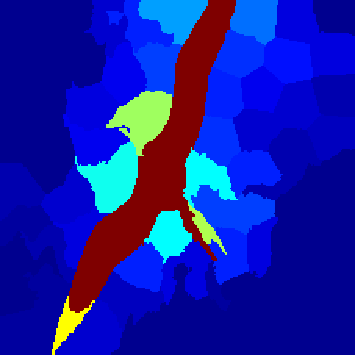}}
\end{minipage}
\hfill
\begin{minipage}{0.23\linewidth}
    \centerline{\includegraphics[width=2.00cm]{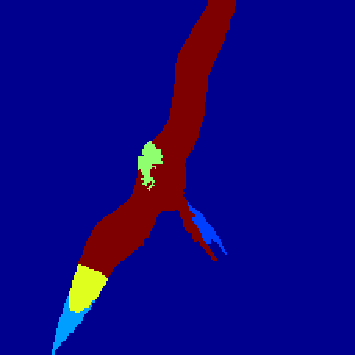}}
\end{minipage}
\vfill

\begin{minipage}{0.23\linewidth}
    \centerline{\includegraphics[width=2.00cm]{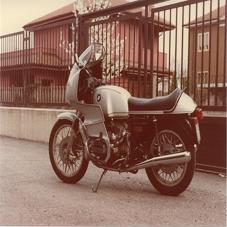}}
\end{minipage}
\hfill
\begin{minipage}{0.23\linewidth}
    \centerline{\includegraphics[width=2.00cm]{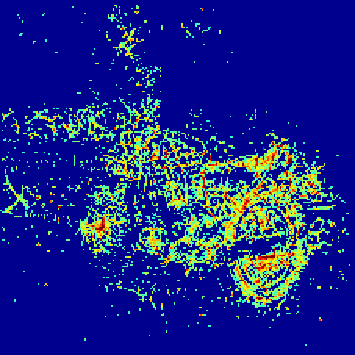}}
\end{minipage}
\hfill
\begin{minipage}{0.23\linewidth}
    \centerline{\includegraphics[width=2.00cm]{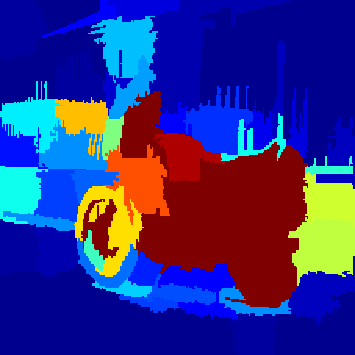}}
\end{minipage}
\hfill
\begin{minipage}{0.23\linewidth}
    \centerline{\includegraphics[width=2.00cm]{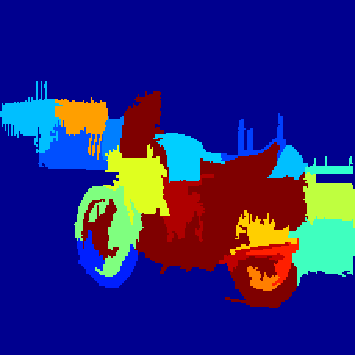}}
\end{minipage}
\vfill

\begin{minipage}{0.23\linewidth}
    \centerline{\includegraphics[width=2.00cm]{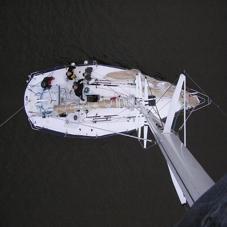}}
\end{minipage}
\hfill
\begin{minipage}{0.23\linewidth}
    \centerline{\includegraphics[width=2.00cm]{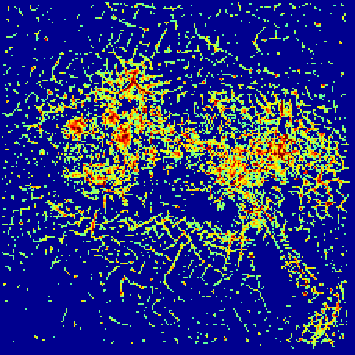}}
\end{minipage}
\hfill
\begin{minipage}{0.23\linewidth}
    \centerline{\includegraphics[width=2.00cm]{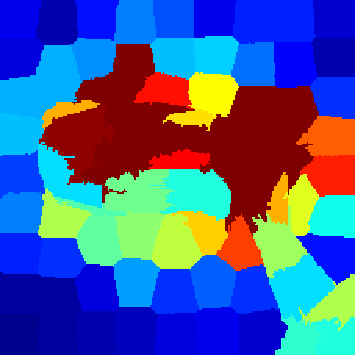}}
\end{minipage}
\hfill
\begin{minipage}{0.23\linewidth}
    \centerline{\includegraphics[width=2.00cm]{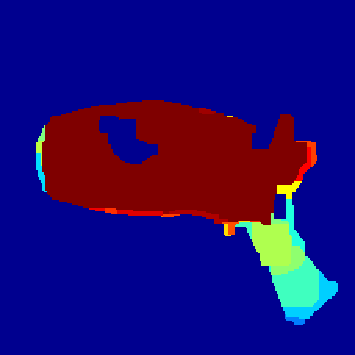}}
\end{minipage}
\vfill

\caption{From left to right: original images, raw saliency maps, smoothed saliency maps and refined saliency maps }
\label{Fig:salMethods}
\end{figure}

At the end of the gradient descent updates, the raw object saliency map ${\bf S}$ is computed as the difference between the modified image and the original one, i.e. $X^{(0)}-X^{(T)}$. For colour images, we average the differences over the RGB channels to obtain a pixel-wise raw saliency map, which is then normalized to be of unit norm. After that, we can apply a simple threshold to filter out some weak signals (in most situations they are corresponding to background) of the raw saliency maps (see the second column in Figure~\ref{Fig:salMethods}).

\subsection{SLIC based saliency map smoothing}
\label{subsec_smoothing}

In practice, we have found that  the continuity of the above raw saliency map ${\bf S}$ is still not good enough in many cases. The main reason is that the DCNN outputs are not totally independent and their correlation is not considered in the above procedure.
Roughly speaking, we have observed that most of the strong signals in the gradients are located in the saliency region. However, from Figure ~\ref{Fig:salMethods} we can see that some problems may still exist, such as background noises, blurred edges or small holes in the foreground. In order to further smooth the saliency maps, we use SLIC superpixels \cite{achanta2012slic} to impose a continuity constraint that all image pixels located in a superpixel always have the same saliency value.  More specifically, we firstly generate the superpixel maps of all test images (In our experiments we will first spilt each test image into $100$ superpixels, and the compact factor is set to $10$). If $i$-th pixel in an image belongs to the $j$th superpixel $P_j$, then the smoothed saliency value can be calculated as Eq.~(\ref{eq-smoothSal}) shows:
\begin{equation}
\label{eq-smoothSal}
\bar{\bf S}_i = \frac{1}{N_j} \sum_{k \in P_j} {\bf S}_k   \;\; (\forall i \in P_j)
\end{equation}
Where $N_j$ is the number of pixels in $P_j$, and we use $\bar{\bf S}$ to denote the smoothed saliency maps. Obviously, comparing with ${\bf S}$, we can see that $\bar{\bf S}$ may fill holes in the saliency regions, sharpen the object edges, and also significantly reduce  the isolated background noises (see the third column in Figure~\ref{Fig:salMethods}).

\subsection{Refine saliency maps using low level features}
\label{subsec_refine}

In Section~\ref{subsec_smoothing}, we have generated the smoothed saliency maps, which can provide much better performance than the original raw saliency maps. On top of that, we  propose to introduce some constraints based on low-level features to further improve the quality of the saliency maps.

Based on the main idea of \cite{fu2013superpixel}, we can generate low level saliency features for each test image. Firstly, we apply the SLIC superpixel generation method in \cite{achanta2012slic} to generate superpixel maps for the test images. Next, for one superpixel $P_i$ in an image, we calculate its color feature $C_i$ by averaging the LAB color value over its all pixels, and use the color feature to calculate its global color contrast $GC_i$ as follows:
\begin{equation}
\label{eq-colorContrast}
GC_i = \sum_{j} \parallel C_i - C_j \parallel_{2}^2.
\end{equation}
where $\parallel \cdot \parallel_2$ denotes the Euclidean distance. Following \cite{fu2013superpixel}, we can further smooth the global color contrast maps and calculate the color distribution maps as the raw low-level saliency maps, which is denoted as  $S_L$. Moreover, $S_L$ is applied to refine the smoothed saliency map $\bar{\bf S}$,  generated from the last step. Here, we normalize $S_L$ between $\alpha$ and $1 + \alpha$, where $0 < \alpha < 1$. The reason to use $\alpha$ is that the low level features contain a lot of errors, which may over-smooth some saliency values in the foreground of some images. By using $\alpha$, we can prevent this refining procedure from removing  some correct saliency regions in $\bar{\bf S}$. The refined saliency map $\hat{\bf S}$ can be generated as:
\begin{equation}\label{refined-saliency}
\hat{\bf S} = S_L \odot \bar{\bf S}.
\end{equation}
where $\odot$ denotes the element-wise multiplication. At the end,  we may further filter out some weak signals in $\hat{S}$ and re-normalize it (see the fourth column in Figure~\ref{Fig:salMethods}). The entire algorithm to generate the final saliency maps is shown in {\bf Algorithm 1}. 

\begin{algorithm}[tb]
   \caption{DCNN based Object Saliency Detection}
   \label{alg:Saliency}
\begin{algorithmic}
   \STATE {\bfseries Input:} an input image $X$, DCNN, SLIC superpixel map $P$, low level saliency feature $S_L$;

   \STATE Use DCNN to recognize the object label for $X$ as $l$;

   \STATE $X^{(0)}=X$;

   \FOR{{\bfseries each} epoch $t=1$ {\bfseries to} $T$}

   \STATE {\bfseries forward pass}: compute the cost function ${\cal F}(X|l)$ ;

   \STATE {\bfseries backward pass}: back-propagate to input layer to compute gradient: $\frac{\partial {\cal F}(X|l)}{\partial X}$;

   \STATE $X^{(t)} \gets X^{(t-1)} - \epsilon \cdot \max\left(\frac{\partial {\cal F}(X|l)}{\partial X},0 \right)$;

   \ENDFOR

   \STATE Average over RGB: ${\bf S}= \frac{1}{3} \sum_{i=1}^3 (X^{(0)}_i - X^{(T)}_i )$;

   \STATE Prune noises with a threshold $\theta$: ${\bf S} = \max({\bf S}-\theta,0)$;

   \STATE Normalize: ${\bf S} = \frac{\bf S}{\| {\bf S} \|}$;

   \STATE Smoothing: using $P$ to smooth ${\bf S}$ as $\bar{\bf S}$;

   \STATE Refine: $\hat{\bf S} = S_L \cdot \bar{\bf S}$;

   \STATE Prune noises and normalize again;

   \STATE {\bfseries Output:} the refined saliency map $\hat{\bf S}$;

\end{algorithmic}
\end{algorithm}

\section{Experiments}
\label{sec_experiments}

We select two benchmark databases to evaluate the performance of the proposed object saliency detection and image segmentation methods, namely Pascal VOC 2012 \cite{Everingham10} and MSRA10k \cite{SalObjSurvey}. For Pascal VOC 2012, we use the $1449$ validation images in its segmentation task as the test set, while for MSRA10k we directly use all $10,000$ images to do the test. Both databases provide the pixel-wise segmentation map (ground truth), thus we can easily measure the performances of different saliency algorithms. Here we compare our approaches with three exisiting methods: i) the first one is the Region Contrast saliency method and the SaliencyCut segmentation method in \cite{cheng2011global}. This method is one of the most popular  bottom-up image saliency detection methods in the literature and it has achieved the state-of-the-art image saliency and segmentation performance on many tasks; ii) the second one is the DCNN based image saliency detection method proposed in \cite{Oxford-cnn-2014}. Similar to our approaches, this method also uses DCNNs and the back-propagation algorithm to generate saliency maps; iii) the third one is the multi-context deep learning based saliency proposed by Zhao et al. \cite{zhao2015saliency} This method uses two DCNNs to calculate global context and local context respectively, and the two level contexts are further combined to generate the final multi-context saliency maps. This method is one of the state-of-the-art deep learning based image saliency algorithm. In our experiments, we use the precision-recall curves (PR-curves) against the ground truth as one metric to evaluate the performance of saliency detection.

As \cite{cheng2011global}, for each saliency map, we vary the cutoff threshold from $0$ to $255$ to generate $256$ precision and recall pairs, which are used to plot a PR-curve. Besides, we also use $F_{\beta}$ to measure the performance for both saliency detection and segmentation, which is  calculated based on precision $Prec$ and recall $Rec$ values with a non-negative weight parameter $\beta$ as follows \cite{borji2012salient}:
\begin{equation}
\label{eq-Fbeta}
F_{\beta} = \frac{(1+{\beta}^2)Prec \times Rec}{{\beta}^2 Prec + Rec}
\end{equation}

In this paper, we follow \cite{cheng2011global} to set ${\beta}^2 = 0.3$ to emphasize the importance of $Prec$. We may derive a sequence of $F_{\beta}$ values along the PR-curve for each saliency map and the largest one is selected as the performance measure (see \cite{borji2012salient}).

\subsection{Databases}

Pascal VOC 2012 database \cite{Everingham10} is a classical image database that can be used for several vision tasks including image classification and saliency. This database currently contains $5717$ training images and $5823$ validation images with $20$ labeled categories. However, among them, only $1449$ validation images that include ground truth information are used to evaluate the performance in our image saliency tasks. Therefore, to expand the training set and improve the classification performance of the DCNN, we merge the original training set with the remaining $4374$ validation images without ground truth to form a new training set, which has $10197$ training samples. For images that are labelled to have more than one class of objects, we use the area of the labelled objects to measure their importance, and use the class of the largest object to label the images for our DCNN training process.

Unfortunately, the Pascal training set is still relatively small for DCNN training. Therefore, we have used a pre-trained DCNN for the ImageNet database, which contains $13$ convolutional layers and $3$ fully connected layers\footnote{We use the net {\em imagenet-vgg-verydeep-16} \cite{simonyan2014vgg}.}. We only use the above-mentioned training data to fine-tune this DCNN for each task with MatConvNet in \cite{MatConvNet-2014}. Here we considered $3$ fine-tune strategies: 1) update the parameters of all hidden layers with same learning rates; 2) update all hidden layers, but only apply large learning rate for the last layer, which corresponding to the output of the DCNN; 3) only update the last layer, and keep other parameters unchange. We have listed top-1 and top-5 classification error rates to measure the performance of the $3$ fine-tune methods. Based on the performance of the $3$ methods, the fine-tuned DCNN from method 1 are used to recognize the test sets on the two tasks we selected.

\begin{table}
\begin{center}
\begin{tabular}{|c|c|c|c|c|c|r|}
\hline
            & Method1     & Method2     & Method3    \\
\hline\hline
Top-1 Err   & 18.0\%      & 20.4\%      & 19.1\%   \\
Top-5 Err   & 1.74\%      & 2.08\%      & 1.79\%   \\
\hline
\end{tabular}
\end{center}
\caption{The classification error rates of three fine-tune methods on the Pascal VOC 2012 test sets.}
\label{table:CNNbaseline}
\end{table}

The classification errors on the test sets imply that the training sample size of Pascal VOC 2012 is still not enough for training deep convolutional networks well. However, as we will see, the proposed algorithms can still yield good performance for saliency detection. If we have more training data, we may expect even better saliency results.

MSRA10k \cite{SalObjSurvey} is another widely-used image saliency database, which is constructed based on Microsoft MSRA saliency database \cite{liu2011learning}. MSRA10k selects $10,000$ images from MSRA and includes pixel-wised salient objects information instead of bounding boxes, which make it suitable for our task. However, MSRA10k dose not include the corresponding training set and class labels of all images. Therefore, for MSRA10k, we directly use the DCNN {\em imagenet-vgg-verydeep-16} \cite{simonyan2014vgg} (without any fine-tuning) to proceed our algorithm.

\subsection{Saliency Results}

In this part we will provide saliency detection results on the selected two databases. In the following,  the PR-curves, $F_{\beta}$ values and some sample images will be used to compare different methods.

\subsubsection{Efficiency}

We firstly consider the speed of our saliency method. Here we will not take the DCNN training time into account because for all of the experiments based on one database, we need only train DCNN once. We can even directly use the will trained DCNN for ImageNet classification for our method without any fine-tune, and the saliency results are also good. Our computing platform includes Intel Xeon E5-1650 CPU (6 cores), 64 GB memory and Nvidia Geforce TITAN X GPU (12 GB memory). The time consumption of processing one image of different algorithms are listed in Table ~\ref{table:time}.

\begin{table}
\begin{center}
\small
\begin{tabular}{|c|c|c|c|c|c|r|}
\hline
 Methods       & RC                     & Method                    & Deep       & Our     \\
               & \cite{cheng2011global} & in \cite{Oxford-cnn-2014} & Saliency \cite{zhao2015saliency} & Method \\
\hline\hline
Execution time   & 1.92s                 & 0.22s                            & 4.38s        & 0.45s \\
\hline
\end{tabular}
\end{center}
\caption{The time for processing one image of different saliency methods.}
\label{table:time}
\end{table}

From Table ~\ref{table:time} we can learn that our method yields much faster processing speed than \cite{cheng2011global} and \cite{zhao2015saliency}. Due to the introducing of SLIC superpixel and low level feature, our method is slower than \cite{Oxford-cnn-2014}. However, in the next part we can find that the proposed method has much better performance than \cite{Oxford-cnn-2014}.

\subsubsection{Pascal VOC 2012}

\begin{figure}[htb]
\begin{center}
    \includegraphics[width=0.85\linewidth]{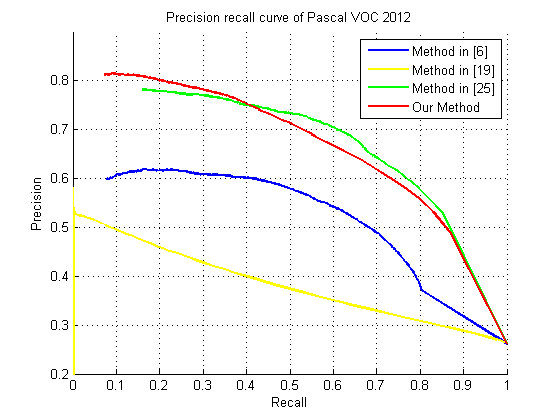}
\end{center}
\caption{The PR-curves of different saliency methods on the Pascal VOC 2012 test set.}
\label{Fig:PRCurve_Pascal}
\end{figure}

For the object saliency detection, we first plot the PR-curves for different methods, which are all shown in Figure~\ref{Fig:PRCurve_Pascal}. From the PR-curves,  we can see that the performance of our proposed saliency detection methods significantly outperform the region contrast in \cite{cheng2011global} and the DCNN based saliency method in \cite{Oxford-cnn-2014}. The proposed method also yield comparable performance as the method in \cite{zhao2015saliency}.

\begin{figure}[htb]
\begin{center}
    \includegraphics[width=0.75\linewidth]{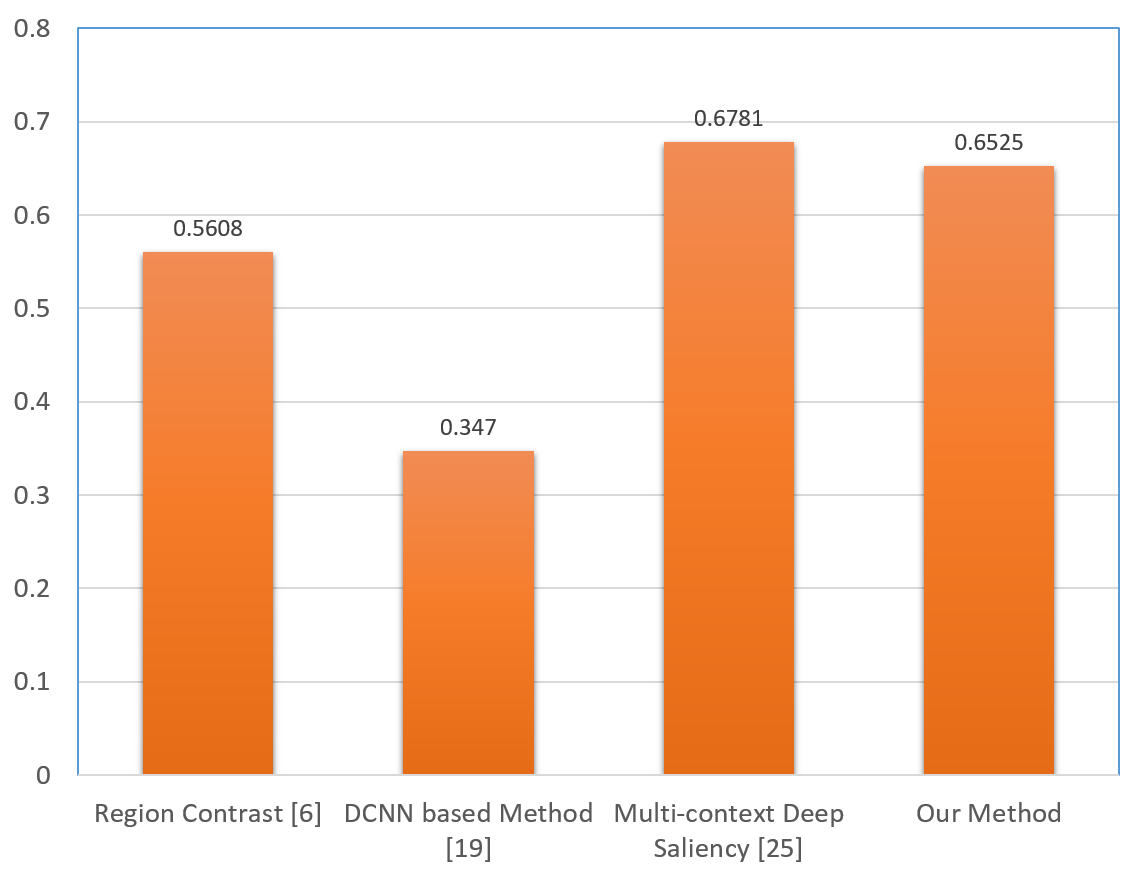}
\end{center}
\caption{The $F_{\beta}$ values of different saliency methods on Pascal VOC 2012 test set.}
\label{Fig:F_Pascal}
\end{figure}

Figure~\ref{Fig:F_Pascal} shows the $F_{\beta}$ values of the different saliency and segmentation methods, from which we can see that the proposed saliency detection method gives the better $F_{\beta}$ value than \cite{cheng2011global} and \cite{Oxford-cnn-2014}, and also similar with \cite{zhao2015saliency}. However, comparing with \cite{zhao2015saliency}, our method yields much faster speed. Finally, in Figure~\ref{Fig:Saliency} (Row 1 to 5), we provide some examples of the saliency detection results from the Pascal VOC 2012 validation set. From these examples we can see that the region contrast algorithm does not work well when the input images have complex background or contain highly variable salient objects, and this problem is fairly common among most bottom-up saliency and segmentation algorithms. On the other hand, we can also see that with the help of SLIC superpixels and low level features, our method can provide comparable performance with \cite{zhao2015saliency}.

\subsubsection{MSRA10k}

\begin{figure}[htb]
\begin{center}
    \includegraphics[width=0.85\linewidth]{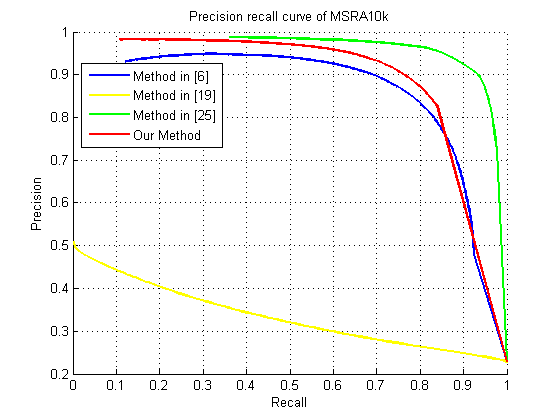}
\end{center}
\caption{The PR-curves of different saliency methods on MSRA10k dataset.}
\label{Fig:PRCurve_MSRA}
\end{figure}

\begin{figure}[htb]
\begin{center}
    \includegraphics[width=0.75\linewidth]{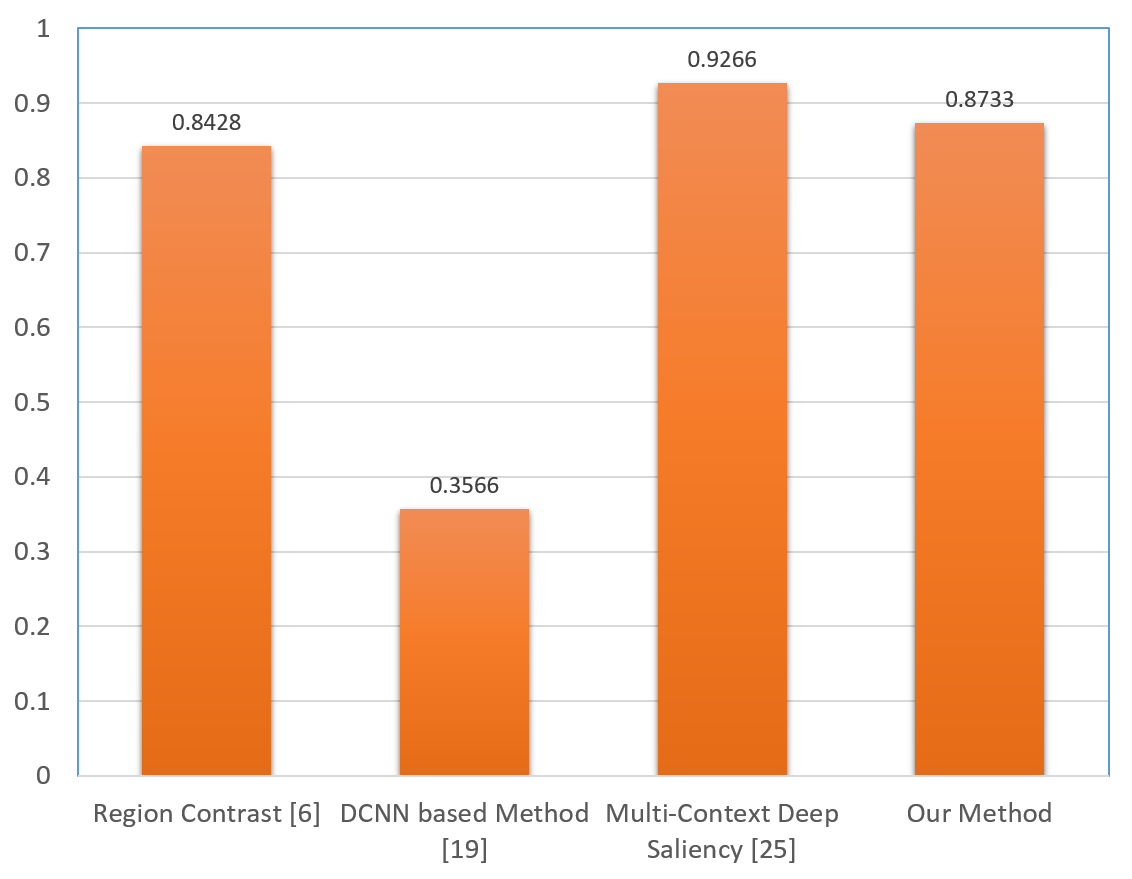}
\end{center}
\caption{The $F_{\beta}$ values of different saliency methods on MSRA10k dataset.}
\label{Fig:F_MSRA}
\end{figure}

Similarly, we also use PR-curves and $F_{\beta}$ to evaluate the saliency and segmentation performance on MSRA10k database. From Figure~\ref{Fig:PRCurve_MSRA}, we can see that the proposed method is significantly better than \cite{Oxford-cnn-2014}, and also has slightly better performance than \cite{cheng2011global}. As shown in Figure~\ref{Fig:F_MSRA}, our methods also give better $F_{\beta}$  value than \cite{cheng2011global} and \cite{Oxford-cnn-2014}.

From Figures ~\ref{Fig:PRCurve_MSRA} and ~\ref{Fig:F_MSRA}, we can see that our method performs slightly worse than \cite{zhao2015saliency} in the MSRA10k dataset. The main reason is attributed to that we directly use a mismatched DCNN trained from the ImageNet dataset. We can not fine-tune the model for this database due to the lack of class labels in MSRA10k. As shown in the figures, the gap between two methods is very small even though we use a mismatched DCNN for our method.

\begin{figure*}[tb]
\begin{minipage}{0.11\linewidth}
    \centerline{\scalebox{0.65}{(A) Original}}
\end{minipage}
\hfill
\begin{minipage}{0.11\linewidth}
    \centerline{\scalebox{0.65}{(B) Ground Truth}}
\end{minipage}
\hfill
\begin{minipage}{0.11\linewidth}
    \centerline{\scalebox{0.65}{(C) RC \cite{cheng2011global}}}
\end{minipage}
\hfill
\begin{minipage}{0.11\linewidth}
    \centerline{\scalebox{0.65}{(D) Method in \cite{Oxford-cnn-2014}}}
\end{minipage}
\hfill
\begin{minipage}{0.11\linewidth}
    \centerline{\scalebox{0.65}{(E) Deep }}
    \centerline{\scalebox{0.65}{ Saliency in \cite{zhao2015saliency}}}
\end{minipage}
\hfill
\begin{minipage}{0.11\linewidth}
    \centerline{\scalebox{0.65}{(F) Our Raw }}
    \centerline{\scalebox{0.65}{Saliency Maps}}
\end{minipage}
\hfill
\begin{minipage}{0.11\linewidth}
    \centerline{\scalebox{0.65}{(G) Our Smoothed}}
    \centerline{\scalebox{0.65}{Saliency Maps}}
\end{minipage}
\hfill
\begin{minipage}{0.11\linewidth}
    \centerline{\scalebox{0.65}{(H) Our Refined}}
    \centerline{\scalebox{0.65}{Saliency Maps}}
\end{minipage}
\vfill

\begin{minipage}{0.12\linewidth}
    \centerline{\includegraphics[width=2.01cm]{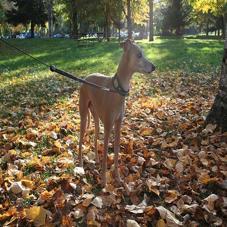}}
\end{minipage}
\hfill
\begin{minipage}{0.12\linewidth}
    \centerline{\includegraphics[width=2.01cm]{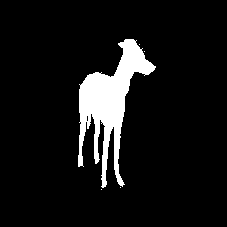}}
\end{minipage}
\hfill
\begin{minipage}{0.12\linewidth}
    \centerline{\includegraphics[width=2.01cm]{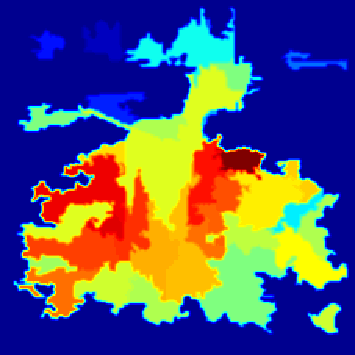}}
\end{minipage}
\hfill
\begin{minipage}{0.12\linewidth}
    \centerline{\includegraphics[width=2.01cm]{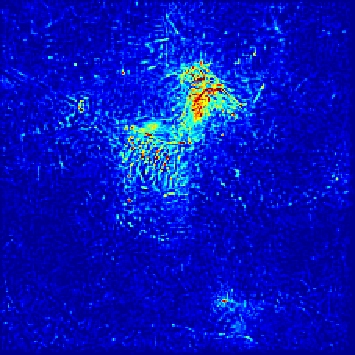}}
\end{minipage}
\hfill
\begin{minipage}{0.12\linewidth}
    \centerline{\includegraphics[width=2.01cm]{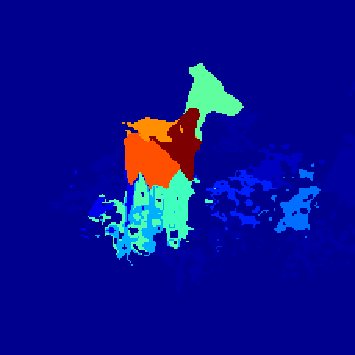}}
\end{minipage}
\hfill
\begin{minipage}{0.12\linewidth}
    \centerline{\includegraphics[width=2.01cm]{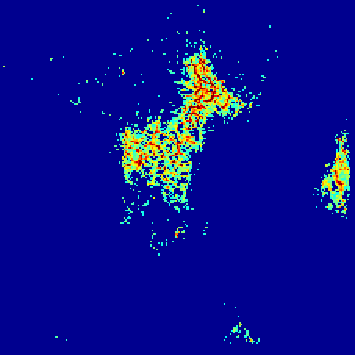}}
\end{minipage}
\hfill
\begin{minipage}{0.12\linewidth}
    \centerline{\includegraphics[width=2.01cm]{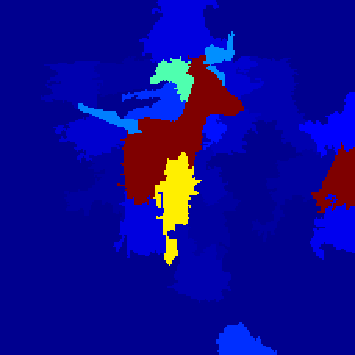}}
\end{minipage}
\hfill
\begin{minipage}{0.12\linewidth}
    \centerline{\includegraphics[width=2.01cm]{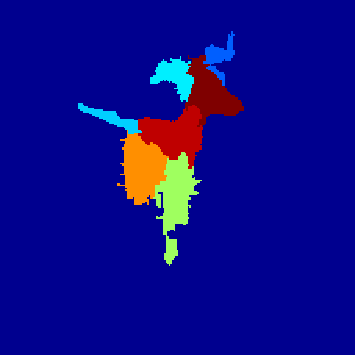}}
\end{minipage}
\vfill

\begin{minipage}{0.12\linewidth}
    \centerline{\includegraphics[width=2.01cm]{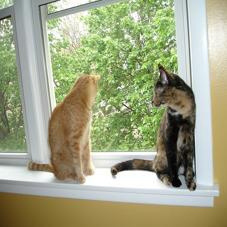}}
\end{minipage}
\hfill
\begin{minipage}{0.12\linewidth}
    \centerline{\includegraphics[width=2.01cm]{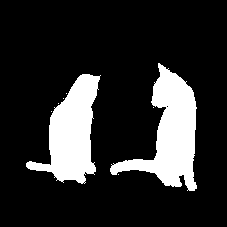}}
\end{minipage}
\hfill
\begin{minipage}{0.12\linewidth}
    \centerline{\includegraphics[width=2.01cm]{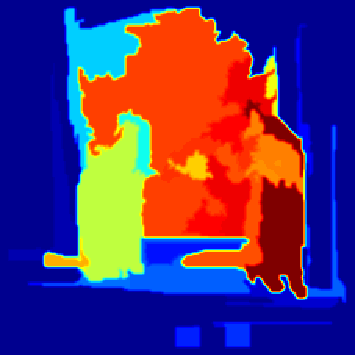}}
\end{minipage}
\hfill
\begin{minipage}{0.12\linewidth}
    \centerline{\includegraphics[width=2.01cm]{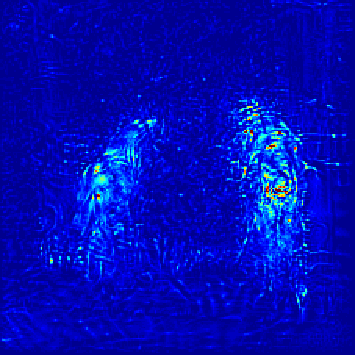}}
\end{minipage}
\hfill
\begin{minipage}{0.12\linewidth}
    \centerline{\includegraphics[width=2.01cm]{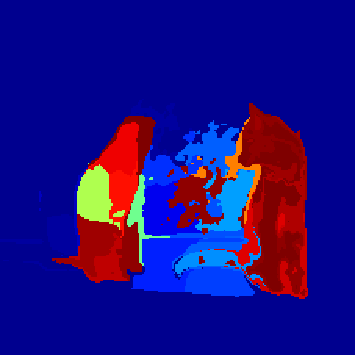}}
\end{minipage}
\hfill
\begin{minipage}{0.12\linewidth}
    \centerline{\includegraphics[width=2.01cm]{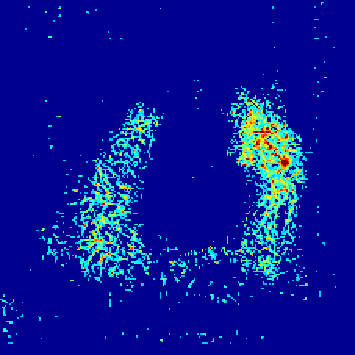}}
\end{minipage}
\hfill
\begin{minipage}{0.12\linewidth}
    \centerline{\includegraphics[width=2.01cm]{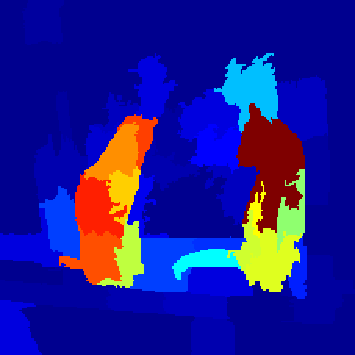}}
\end{minipage}
\hfill
\begin{minipage}{0.12\linewidth}
    \centerline{\includegraphics[width=2.01cm]{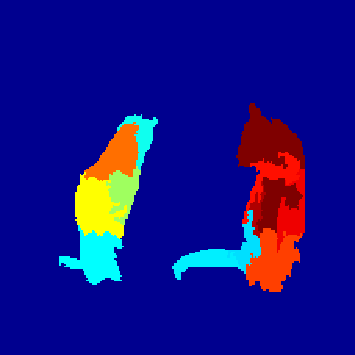}}
\end{minipage}
\vfill

\begin{minipage}{0.12\linewidth}
    \centerline{\includegraphics[width=2.01cm]{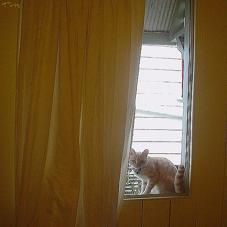}}
\end{minipage}
\hfill
\begin{minipage}{0.12\linewidth}
    \centerline{\includegraphics[width=2.01cm]{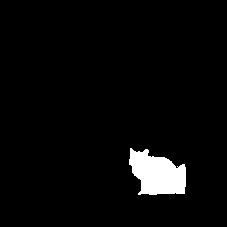}}
\end{minipage}
\hfill
\begin{minipage}{0.12\linewidth}
    \centerline{\includegraphics[width=2.01cm]{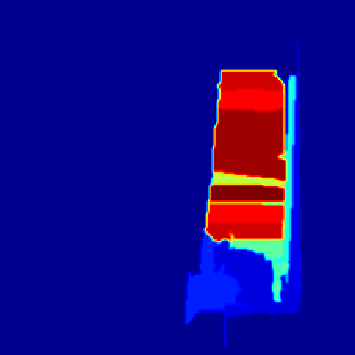}}
\end{minipage}
\hfill
\begin{minipage}{0.12\linewidth}
    \centerline{\includegraphics[width=2.01cm]{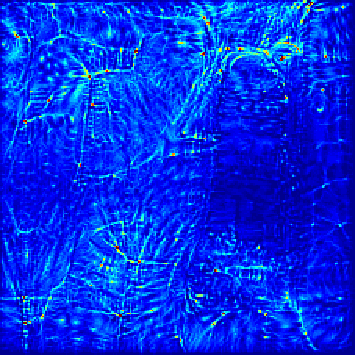}}
\end{minipage}
\hfill
\begin{minipage}{0.12\linewidth}
    \centerline{\includegraphics[width=2.01cm]{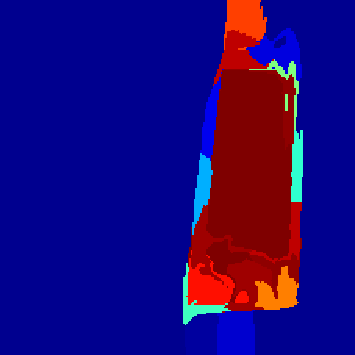}}
\end{minipage}
\hfill
\begin{minipage}{0.12\linewidth}
    \centerline{\includegraphics[width=2.01cm]{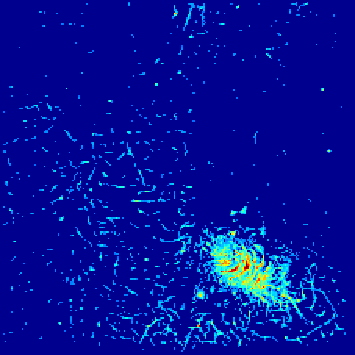}}
\end{minipage}
\hfill
\begin{minipage}{0.12\linewidth}
    \centerline{\includegraphics[width=2.01cm]{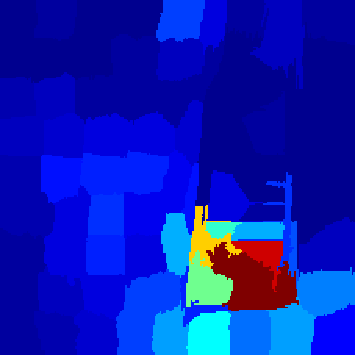}}
\end{minipage}
\hfill
\begin{minipage}{0.12\linewidth}
    \centerline{\includegraphics[width=2.01cm]{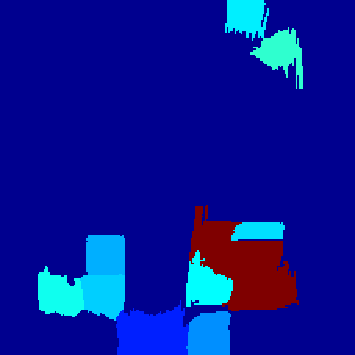}}
\end{minipage}
\vfill

\begin{minipage}{0.12\linewidth}
    \centerline{\includegraphics[width=2.01cm]{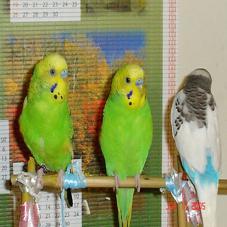}}
\end{minipage}
\hfill
\begin{minipage}{0.12\linewidth}
    \centerline{\includegraphics[width=2.01cm]{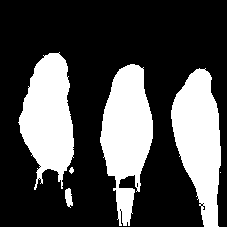}}
\end{minipage}
\hfill
\begin{minipage}{0.12\linewidth}
    \centerline{\includegraphics[width=2.01cm]{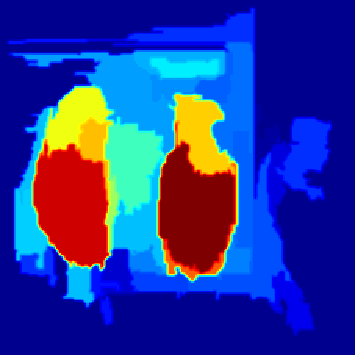}}
\end{minipage}
\hfill
\begin{minipage}{0.12\linewidth}
    \centerline{\includegraphics[width=2.01cm]{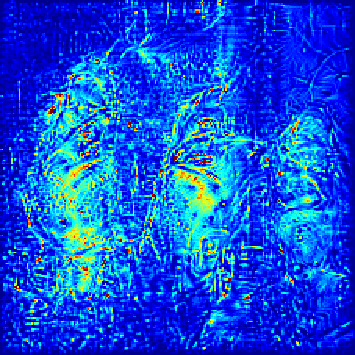}}
\end{minipage}
\hfill
\begin{minipage}{0.12\linewidth}
    \centerline{\includegraphics[width=2.01cm]{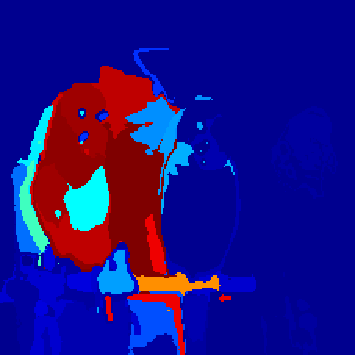}}
\end{minipage}
\hfill
\begin{minipage}{0.12\linewidth}
    \centerline{\includegraphics[width=2.01cm]{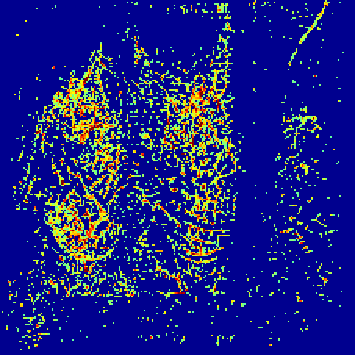}}
\end{minipage}
\hfill
\begin{minipage}{0.12\linewidth}
    \centerline{\includegraphics[width=2.01cm]{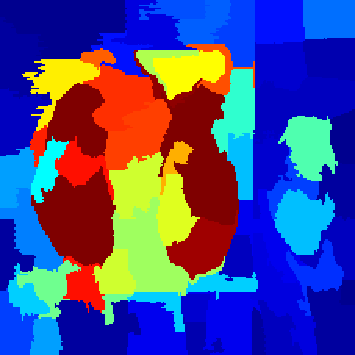}}
\end{minipage}
\hfill
\begin{minipage}{0.12\linewidth}
    \centerline{\includegraphics[width=2.01cm]{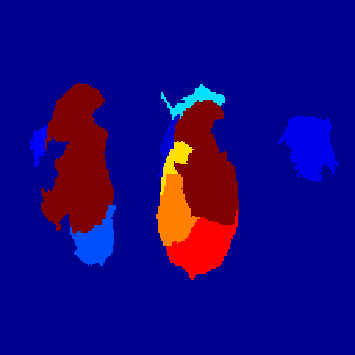}}
\end{minipage}
\vfill

\begin{minipage}{0.12\linewidth}
    \centerline{\includegraphics[width=2.01cm]{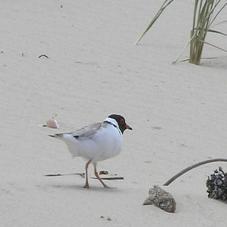}}
\end{minipage}
\hfill
\begin{minipage}{0.12\linewidth}
    \centerline{\includegraphics[width=2.01cm]{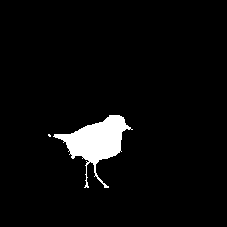}}
\end{minipage}
\hfill
\begin{minipage}{0.12\linewidth}
    \centerline{\includegraphics[width=2.01cm]{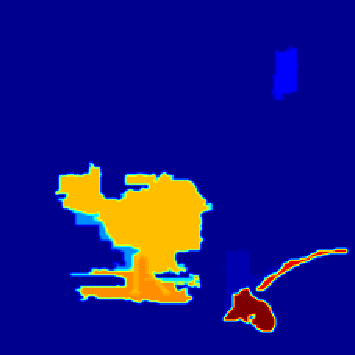}}
\end{minipage}
\hfill
\begin{minipage}{0.12\linewidth}
    \centerline{\includegraphics[width=2.01cm]{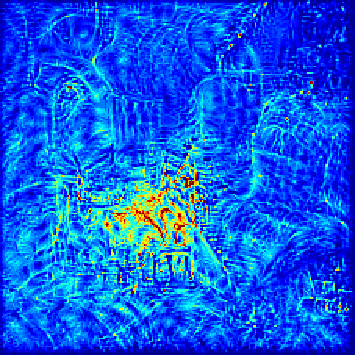}}
\end{minipage}
\hfill
\begin{minipage}{0.12\linewidth}
    \centerline{\includegraphics[width=2.01cm]{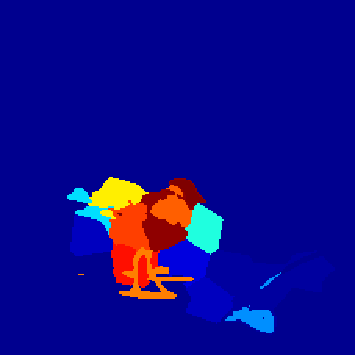}}
\end{minipage}
\hfill
\begin{minipage}{0.12\linewidth}
    \centerline{\includegraphics[width=2.01cm]{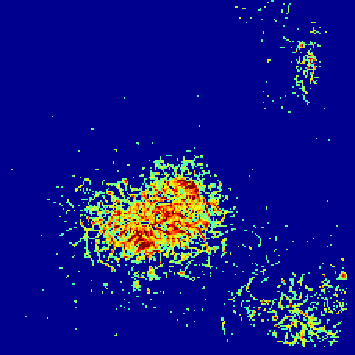}}
\end{minipage}
\hfill
\begin{minipage}{0.12\linewidth}
    \centerline{\includegraphics[width=2.01cm]{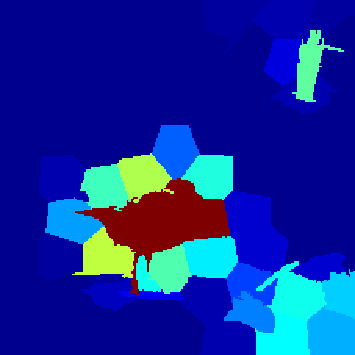}}
\end{minipage}
\hfill
\begin{minipage}{0.12\linewidth}
    \centerline{\includegraphics[width=2.01cm]{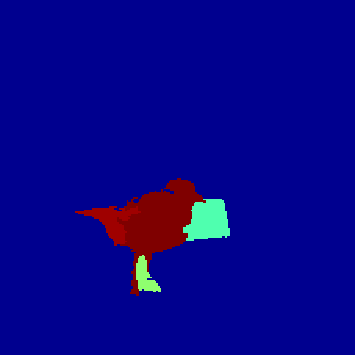}}
\end{minipage}
\vfill

\begin{minipage}{0.12\linewidth}
    \centerline{\includegraphics[width=2.01cm]{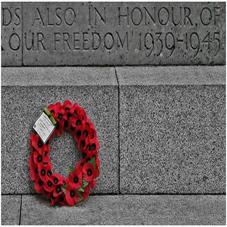}}
\end{minipage}
\hfill
\begin{minipage}{0.12\linewidth}
    \centerline{\includegraphics[width=2.01cm]{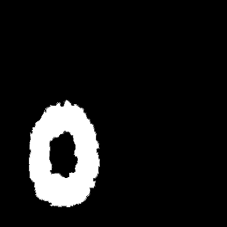}}
\end{minipage}
\hfill
\begin{minipage}{0.12\linewidth}
    \centerline{\includegraphics[width=2.01cm]{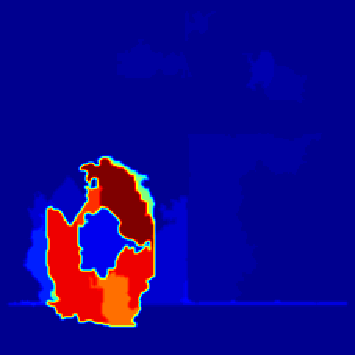}}
\end{minipage}
\hfill
\begin{minipage}{0.12\linewidth}
    \centerline{\includegraphics[width=2.01cm]{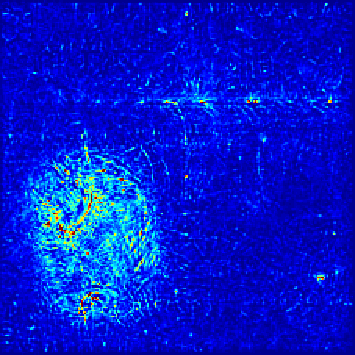}}
\end{minipage}
\hfill
\begin{minipage}{0.12\linewidth}
    \centerline{\includegraphics[width=2.01cm]{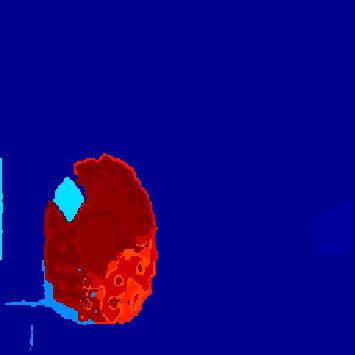}}
\end{minipage}
\hfill
\begin{minipage}{0.12\linewidth}
    \centerline{\includegraphics[width=2.01cm]{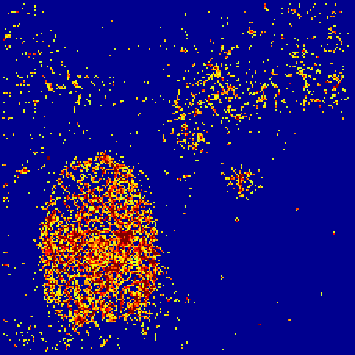}}
\end{minipage}
\hfill
\begin{minipage}{0.12\linewidth}
    \centerline{\includegraphics[width=2.01cm]{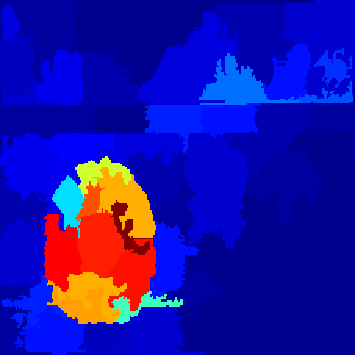}}
\end{minipage}
\hfill
\begin{minipage}{0.12\linewidth}
    \centerline{\includegraphics[width=2.01cm]{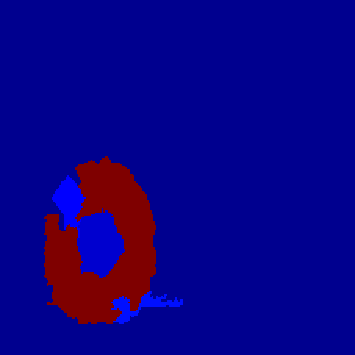}}
\end{minipage}
\vfill

\begin{minipage}{0.12\linewidth}
    \centerline{\includegraphics[width=2.01cm]{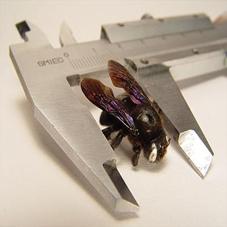}}
\end{minipage}
\hfill
\begin{minipage}{0.12\linewidth}
    \centerline{\includegraphics[width=2.01cm]{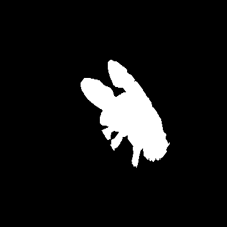}}
\end{minipage}
\hfill
\begin{minipage}{0.12\linewidth}
    \centerline{\includegraphics[width=2.01cm]{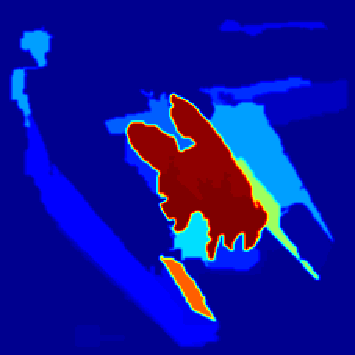}}
\end{minipage}
\hfill
\begin{minipage}{0.12\linewidth}
    \centerline{\includegraphics[width=2.01cm]{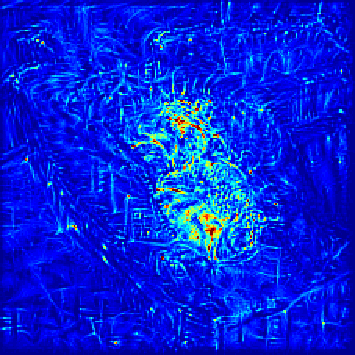}}
\end{minipage}
\hfill
\begin{minipage}{0.12\linewidth}
    \centerline{\includegraphics[width=2.01cm]{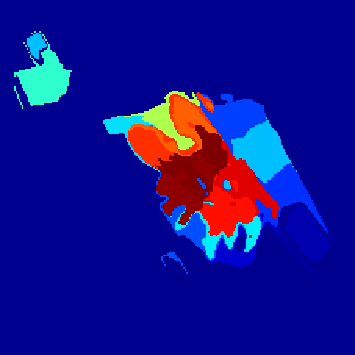}}
\end{minipage}
\hfill
\begin{minipage}{0.12\linewidth}
    \centerline{\includegraphics[width=2.01cm]{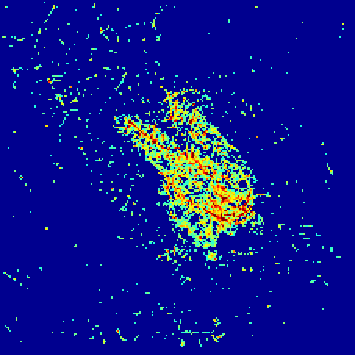}}
\end{minipage}
\hfill
\begin{minipage}{0.12\linewidth}
    \centerline{\includegraphics[width=2.01cm]{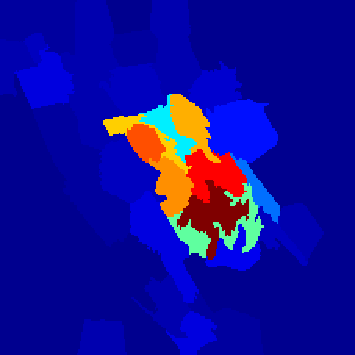}}
\end{minipage}
\hfill
\begin{minipage}{0.12\linewidth}
    \centerline{\includegraphics[width=2.01cm]{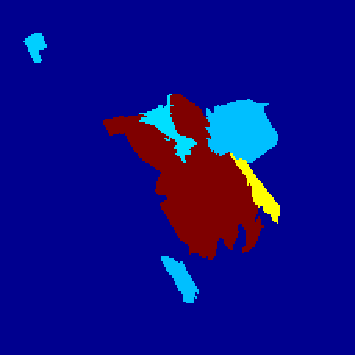}}
\end{minipage}
\vfill

\begin{minipage}{0.12\linewidth}
    \centerline{\includegraphics[width=2.01cm]{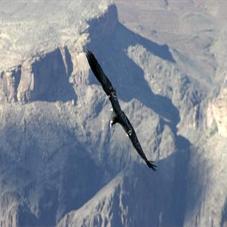}}
\end{minipage}
\hfill
\begin{minipage}{0.12\linewidth}
    \centerline{\includegraphics[width=2.01cm]{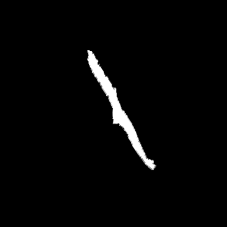}}
\end{minipage}
\hfill
\begin{minipage}{0.12\linewidth}
    \centerline{\includegraphics[width=2.01cm]{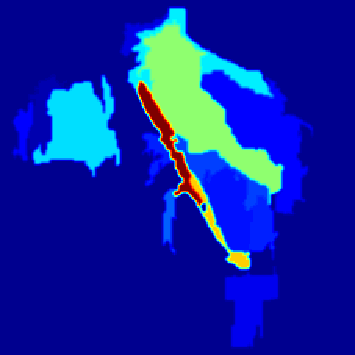}}
\end{minipage}
\hfill
\begin{minipage}{0.12\linewidth}
    \centerline{\includegraphics[width=2.01cm]{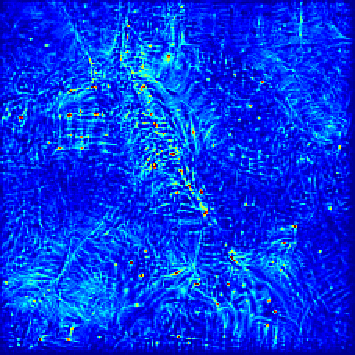}}
\end{minipage}
\hfill
\begin{minipage}{0.12\linewidth}
    \centerline{\includegraphics[width=2.01cm]{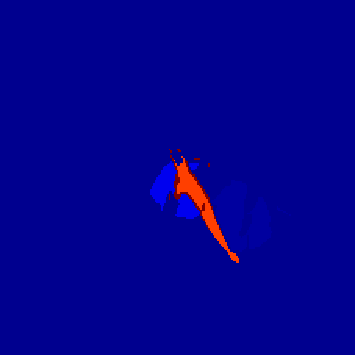}}
\end{minipage}
\hfill
\begin{minipage}{0.12\linewidth}
    \centerline{\includegraphics[width=2.01cm]{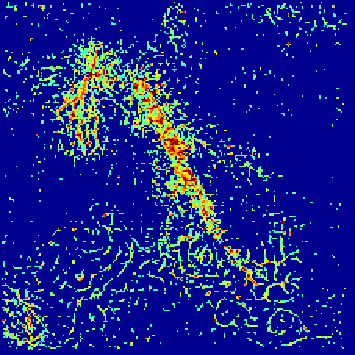}}
\end{minipage}
\hfill
\begin{minipage}{0.12\linewidth}
    \centerline{\includegraphics[width=2.01cm]{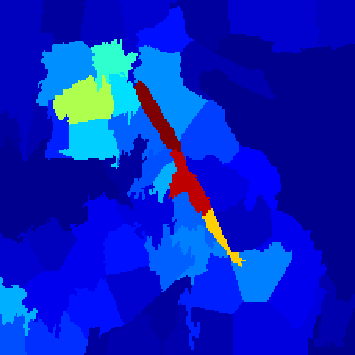}}
\end{minipage}
\hfill
\begin{minipage}{0.12\linewidth}
    \centerline{\includegraphics[width=2.01cm]{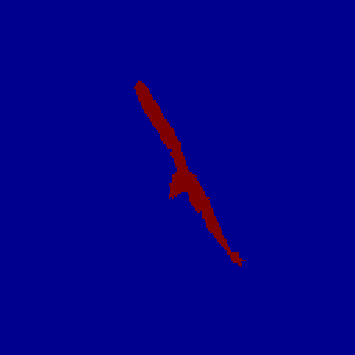}}
\end{minipage}
\vfill

\begin{minipage}{0.12\linewidth}
    \centerline{\includegraphics[width=2.01cm]{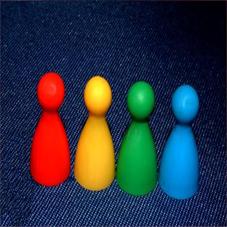}}
\end{minipage}
\hfill
\begin{minipage}{0.12\linewidth}
    \centerline{\includegraphics[width=2.01cm]{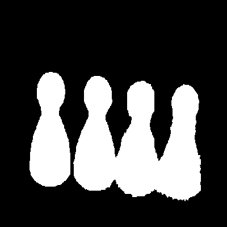}}
\end{minipage}
\hfill
\begin{minipage}{0.12\linewidth}
    \centerline{\includegraphics[width=2.01cm]{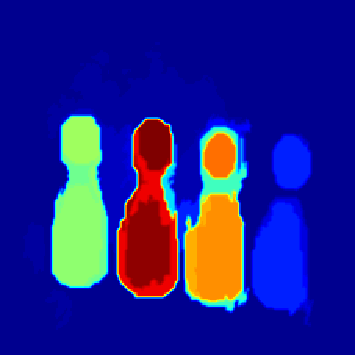}}
\end{minipage}
\hfill
\begin{minipage}{0.12\linewidth}
    \centerline{\includegraphics[width=2.01cm]{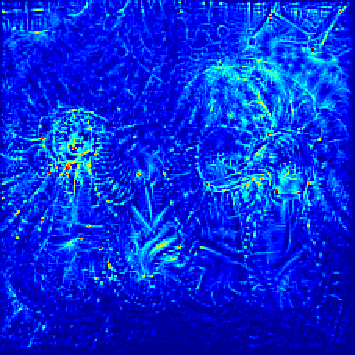}}
\end{minipage}
\hfill
\begin{minipage}{0.12\linewidth}
    \centerline{\includegraphics[width=2.01cm]{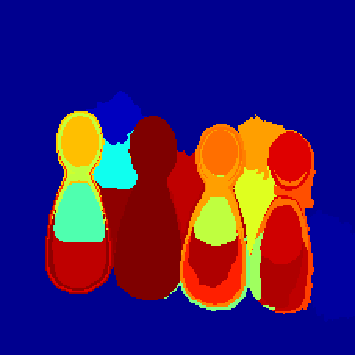}}
\end{minipage}
\hfill
\begin{minipage}{0.12\linewidth}
    \centerline{\includegraphics[width=2.01cm]{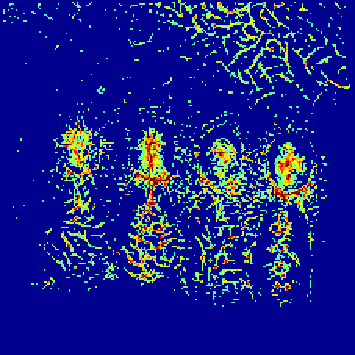}}
\end{minipage}
\hfill
\begin{minipage}{0.12\linewidth}
    \centerline{\includegraphics[width=2.01cm]{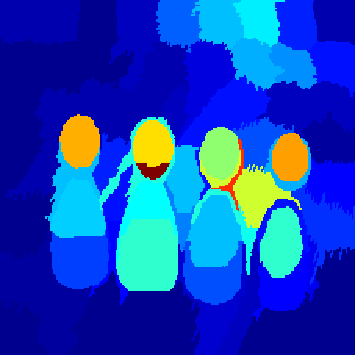}}
\end{minipage}
\hfill
\begin{minipage}{0.12\linewidth}
    \centerline{\includegraphics[width=2.01cm]{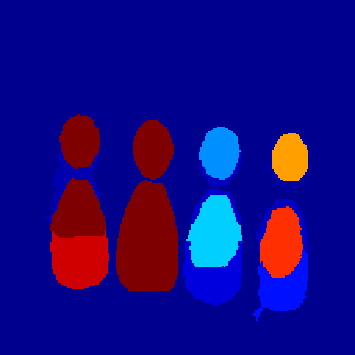}}
\end{minipage}
\vfill

\begin{minipage}{0.12\linewidth}
    \centerline{\includegraphics[width=2.01cm]{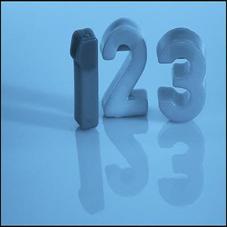}}
\end{minipage}
\hfill
\begin{minipage}{0.12\linewidth}
    \centerline{\includegraphics[width=2.01cm]{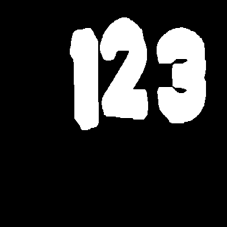}}
\end{minipage}
\hfill
\begin{minipage}{0.12\linewidth}
    \centerline{\includegraphics[width=2.01cm]{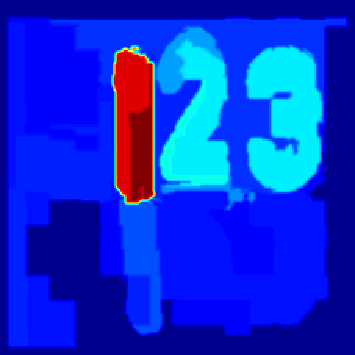}}
\end{minipage}
\hfill
\begin{minipage}{0.12\linewidth}
    \centerline{\includegraphics[width=2.01cm]{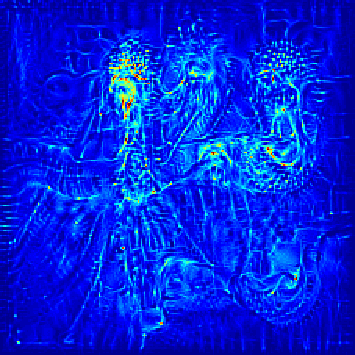}}
\end{minipage}
\hfill
\begin{minipage}{0.12\linewidth}
    \centerline{\includegraphics[width=2.01cm]{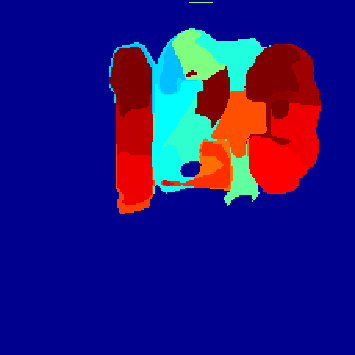}}
\end{minipage}
\hfill
\begin{minipage}{0.12\linewidth}
    \centerline{\includegraphics[width=2.01cm]{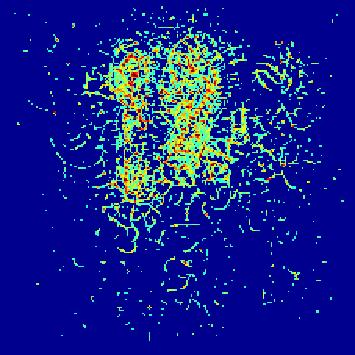}}
\end{minipage}
\hfill
\begin{minipage}{0.12\linewidth}
    \centerline{\includegraphics[width=2.01cm]{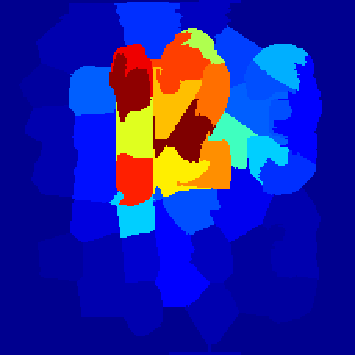}}
\end{minipage}
\hfill
\begin{minipage}{0.12\linewidth}
    \centerline{\includegraphics[width=2.01cm]{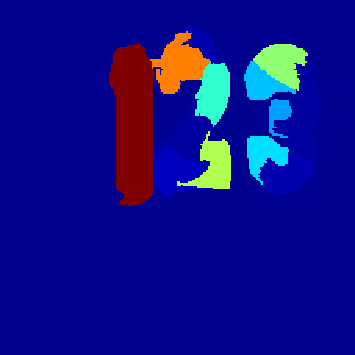}}
\end{minipage}
\vfill

\caption{Saliency Results of Pascal VOC 2012 (Row 1 to 5) and MSRA10k (Row 6 to 10). (A) original images, (B) ground truth, (C) Region Contrast saliency maps \cite{cheng2011global}, (D) DCNN based saliency maps by using \cite{Oxford-cnn-2014}, (E) multi-context deep saliency method \cite{zhao2015saliency}, (F) our raw saliency maps, (G) our smoothed saliency maps, (H) our refined saliency maps. }
\label{Fig:Saliency}
\end{figure*}

In Figure~\ref{Fig:Saliency}, we also select several MSRA10k images to show the saliency results (Row 6 to 10).

\section{Conclusion}
\label{sec_conc}
In this paper, we have proposed a novel DCNN-based method for object saliency detection. The method firstly train a regular DCNN for saliency detection. After that, for each test image, we firstly recognize the image class label, and then we can use the pre-trained DCNN to generate a saliency map. Specifically, we attempt to reduce a cost function defined to measure the class-specific objectness of each image, and we back-propagate the corresponding error signal all way to the input layer and use the gradient of inputs to revise the input images. After several iterations, the difference between the original input images and the revised images is calculated as a raw saliency map. The raw saliency maps are then smoothed and refined by using SLIC superpixels and low level saliency features. We have evaluated our methods on two benchmark tasks, namely Pascal VOC 2012 \cite{Everingham10} and MSRA10k \cite{SalObjSurvey}.
Experimental results  have shown that our proposed methods can generate high-quality saliency maps in relatively short time (nearly 10 times faster than the state-of-the-art DCNN based method in \cite{zhao2015saliency}), which clearly outperforming many other existing methods. Comparing with many low-level feature methods, our DCNN-based approach excels on many difficult images, containing complex background, highly-variable salient objects, multiple objects, and very small objects.


{\small
\bibliography{CNN_Saliency}
}

\end{document}